\documentclass[lettersize,journal]{IEEEtran}

\usepackage{amsmath,amsfonts}
\usepackage{color}
\usepackage{soul}
\usepackage{amssymb}
\usepackage{algorithmic}
\usepackage{algorithm}
\usepackage{array}
\usepackage[caption=false,font=normalsize,labelfont=sf,textfont=sf]{subfig}
\usepackage{textcomp}
\usepackage{stfloats}
\usepackage{url}
\usepackage{verbatim}
\usepackage{graphicx}
\usepackage{cite}
\usepackage{xcolor}
\usepackage{bbding}

\usepackage{hyperref}

\usepackage[acronym]{glossaries}
\makeglossaries

\newacronym{ai}{AI}{Artificial Intelligence}
\newacronym{ml}{ML}{Machine Learning}
\newacronym{mlops}{MLOps}{Machine Learning Operations}
\newacronym{6g}{6G}{6th Generation Wireless Cellular Networks}
\newacronym{sprt}{SPRT}{Sequential Probability Ratio Test}
\newacronym{cumsum}{CUMSUM}{Cumulative Sum}
\newacronym{psi}{PSI}{Population Stability Index}
\newacronym{klt}{KL}{Kullback-Leibler}
\newacronym{pht}{PH}{Page-Hinkley}
\newacronym{adwin}{ADWIN}{ADaptive WINdowing}
\newacronym{edist}{ED}{Energy Distance}
\newacronym{emd}{EMD}{Earth Mover's Distance}
\newacronym{kst}{KS}{Kolmogorov–Smirnov}
\newacronym{mlp}{MLP}{Multi-layer Perceptron}
\newacronym{cnn}{CNN}{Convolutional Neural Network}
\newacronym{pdf}{PDF}{Probability Density Function}
\newacronym{adv}{ADV}{Adversarial Detection}
\newacronym{edc}{EDC}{Embedding-based domain classification}
\newacronym{dde}{DDE}{Data Driven Deep Density Estimation}

\begin{document}

\title{A Representation Learning Approach to Feature Drift Detection in Wireless Networks}

\author{
	\IEEEauthorblockN{
		Athanasios Tziouvaras\IEEEauthorrefmark{1},
		Bla\v{z} Bertalani\v{c}\IEEEauthorrefmark{2},
		George Floros\IEEEauthorrefmark{6}\IEEEauthorrefmark{7},
		Kostas Kolomvatsos\IEEEauthorrefmark{4},
		Panagiotis Sarigiannidis\IEEEauthorrefmark{3},
		and Carolina Fortuna\IEEEauthorrefmark{2}
	}
	
	\IEEEauthorblockA{
		\IEEEauthorrefmark{1} Business and IoT Integrated Solutions LTD, Nicosia, Cyprus\\
		\IEEEauthorrefmark{2} Jo\v{z}ef Stefan Institute, Slovenia\\
		\IEEEauthorrefmark{6} Trinity College Dublin, Dublin, Ireland\\
		\IEEEauthorrefmark{7} University of Thessaly, Volos, Greece\\
		\IEEEauthorrefmark{4} University of Thessaly, Lamia, Greece\\
		\IEEEauthorrefmark{3} University of Western Macedonia, Kozani, Greece\\
		Email: attziouv@bi2s.eu, blaz.bertalanic@ijs.si, florosg@tcd.ie, kostasks@uth.gr, psarigiannidis@uowm.gr, carolina.fortuna@ijs.si
	}
}

\maketitle
\begin{abstract}
\gls{ai} is foreseen to be a centerpiece in next generation wireless networks enabling  ubiquitous communication as well as new services.
However, in real deployment, feature distribution changes may degrade the performance of \gls{ai} models and lead to undesired behaviors. 
To counter for undetected model degradation, we propose ALERT; a method that can detect feature distribution changes and trigger model re-training that works well on two wireless network use cases: wireless fingerprinting and link anomaly detection. ALERT includes three components: representation learning,  statistical testing and  utility assessment. We rely on \gls{mlp} for designing the representation learning component, on \gls{kst} and \gls{psi} tests for designing the statistical testing and a new function for utility assessment. We show the superiority of the proposed method against ten standard drift detection methods available in the literature on two wireless network use cases.
\end{abstract}


\section{Introduction}
\label{sec:intro}

Artificial Intelligence (\gls{ai}) is foreseen to be a centerpiece in next generation wireless networks, including \gls{6g} and beyond cellular networks~\cite{Letaief2019TheNetworks} by enabling ubiquitous communication, new services including high-accuracy localization \cite{10287134}, anomaly fault and detection \cite{9715175} as well as replacing traditionally networking functionality by \gls{ai} based realization towards so-called AI native functionality~\cite{hoydis2021toward,wu2022ai}. \gls{ai} models are typically developed offline by using a pre-defined amount of data and a set of \gls{ml} techniques that are tuned (semi-)manually to find the best performing combination for the respective training data \cite{cabrera2023did}. However, when deployed in a real, production environment, the model development workflow is managed by the so-called \gls{ai}/\gls{ml} workflow~\cite{b1} realized through \gls{mlops} tools ~\cite{cop2025overviewsolutiondemocratizingai}. The combination of those tools and their deployment enable \gls{mlops} pipelines that automatically manage the data preparation, model training, evaluation, selection  and serving in production systems.   

Once deployed in a production setting, the data collection, \gls{mlops} pipeline and integrated \gls{ai} models are typically managed by different teams, sometimes without significant coordination between each other \cite{sculley2015hidden}. In such setting, it may happen that input data distribution changes occur naturally due to changes in the observed systems, but it may also happen that unstable data dependencies such as re-calibrations done by the team responsible for data collection is not propagated to the teams managing the \gls{mlops} or \gls{ai} systems. Therefore, the performance in production degrades and adjustments are reactive rather than proactive \cite{sculley2015hidden}. A recent study across several hundreds of eNodeBs and three categories of wireless KPIs such as resource utilization has also confirmed the existence of drifts in cellular networks \cite{liu2023leaf} while \cite{manias2023model} identified the challenges surrounding the implementation of drift detection and mitigation schemes
in resource-constrained networks.

To detect and signal distribution changes that may degrade the performance of \gls{ai} in production, libraries able to detect feature drifts while integrating with existing \gls{mlops} tools have been developed \cite{swathi2023deployment}. These libraries incorporate several \textit{drift detectors}, defined as methods that observe a stream of data over time and determine for every new data point if the current distribution of the data has changed compared to a reference data set \cite{simonetto2024impact}. Several drift detection techniques as part of three such libraries have been recently benchmarked on two use-cases: occupancy detection and prediction of energy consumption \cite{muller2024open}. To date, the only investigations of the drift phenomenon on wireless data are available in \cite{liu2023leaf} including a Kolmogorov-Smirnov based detection technique and \cite{manias2023model} that considered Isolation Forests and threshold to detect drifts in an illustrative example.

Aiming to provide a better insight into the suitability of existing drift detection techniques on wireless data as well as improve the existing state of the art for detection in wireless networks, we propose a new feature drift detection method (named ALERT), and benchmark it against ten standard  methods on two use cases: wireless fingerprinting and link fault or anomaly detection. The contributions of this paper are:

\begin{itemize}
    \item ALERT, a new feature drift detection method consisting of three components: representation learning,  statistical testing and  utility assessment.
    \item Validation on two wireless Use Cases that utilize real-world data. We show that the ALERT method outperforms all the baseline models, achieving an overall F1-score of $0.9$ in the \textit{fingerprinting} use case and  $0.88$ in the \textit{links} use case.
    \item Analysis (i) identifying feature drift; (ii) assessing their impact on the model; (iii) attempting to answer "when" to retrain the model with the new data.
\end{itemize}

This paper is organized as follows. Section \ref{sec:related} summarizes related work, Section \ref{sec:drift} provides background related to drift detection, Section \ref{sec:statement} provides the problem statement while Section \ref{sec:data_drift_method} introduces the proposed ALERT method. Section \ref{sec:eval} details the evaluation methodology Section \ref{sec:results} while Section \ref{sec:conclusion} concludes the paper. 

\section{Related work}
\label{sec:related}

Data and concept drift are sometimes interchangeably used in the literature while in some cases one is considered as a superset of the other. In this paper we follow the definition from \cite{moreno2012unifying} where \textit{feature drifts} are categorized into four primary types: \textit{covariate drift} also referred to a feature drift in this paper, \textit{prior probability drift}, \textit{concept drift}, and \textit{dataset shift}. We group related works in three categories: works that develop new or analyze existing drift techniques, works that develop drift detection tools and systems and works that focus on studying specific use-cases.

\subsection{Drift Detection Techniques}
\label{sec:drift_tech}
Most of the drift, or change, detection techniques can classified as follows based on the type of performed analysis: 1) sequential analysis, 2) control charts, 3) difference between distributions and 4) contextual as discussed in \cite{gama2014survey} with \cite{1} providing a different grouping.  One of the foundational sequential analysis test is the \gls{sprt} that detects at a point $p$ a change from a distribution to another. Other tests such as \gls{cumsum} use principles from \gls{sprt}. The \gls{pht} test is a sequential adaptation of the detection of an abrupt change in
the average of a Gaussian signal \cite{gama2014survey}. The methods from the second category are based on based on statistical process control represented by standard statistical techniques to monitor and control the quality
of a product during a continuous manufacturing. One such example is the exponentially weighted moving average (EWMA) \cite{gama2014survey}.

The methods from the third group, that monitor distributions on two different time-windows, compare the two distributions computed over the two windows using statistical tests and signal a drift when the distributions are not equal. The \gls{klt} divergence test, the Population Stability Index (\gls{psi}), a variation of the  \gls{klt} \cite{8}, as well as \gls{adwin} all fall under this category \cite{gama2014survey}. Some other statistical tests as follow  probably also fall in this group. The \gls{edist} \cite{harald1928composition} that computes a statistical distance between two probability distributions, the \gls{emd} \cite{rubner2000earth} that computes the minimal cost that must be paid to transform one distribution into the other, the Kolmogorov-Smirnov test (\gls{kst}) the quantifies the  distance between the empirical distribution function of the sample and the cumulative distribution function of the reference distribution, or between the empirical distribution functions of two samples, the Kuiper test that is closely related to \gls{kst}, etc \cite{mayaki2022autoregressive}.

Finally, the contextual detectors rely on learning, with examples such as the Splice that is a meta-
learning technique that implements a context sensitive batch learning approach and the Incremental Fuzzy Classification System algorithm.

\subsection{Drift Detection Tools and Systems}

The authors in \cite{muller2024open} proposed D3Bench, a benchmarking tool that enables the functional and non-functional evaluation of drift detection tools. In their analysis they benchmark three open source tools for drift detection and found that Evidently AI stands out for its general feature drift detection,
whereas NannyML excels at pinpointing the precise timing
of shifts and evaluating their consequent effects on predictive
accuracy. The authors of \cite{dong2024efficiently} start from the observation that observe
that not all feature drifts lead to degradation in prediction accuracy and propose a new strategy which, using decision trees, is able to precisely
pinpoint low-accuracy zones within ML models. The work triggers model improvement through active learning only in cases of harmful drifts that detrimentally affect model performance. Rather than triggering model retraining when drift is noticed, 
\cite{mallick2022matchmaker} proposed Matchmaker, a tool that dynamically identifies the batch of training data that is most
similar to each test sample, and uses the ML model trained
on that data for inference.

\subsection{Drift Detection Use Cases}

The authors in \cite{muller2024open} benchmarked three drift detection tools on univariate data falling under two use-cases: occupancy detection where CO2 and room temperature were used as features while occupancy was the target variable and energy consumption prediction where energy consumption was the feature. The authors of \cite{3} study the impact of industrial delays when mitigating distribution drifts on a financial use-case. Focusing on cellular wireless data,  \cite{liu2023leaf} introduce a methodology for concept drift mitigation that explains the features and time intervals that contribute the most to drift; and mitigates it using forgetting and over-sampling. An illustrative demand prediction use case for multimedia service in a 5G network was briefly considered in \cite{manias2023model}. Isolation Forests and threshold were used to conceptually illustrate  drifts detection.

\section{Drift Definition }
\label{sec:drift}

As briefly mentioned in Section \ref{sec:related}, the terminology related to data, concept and model drift varies across works. The formal mathematical definitions are generally similar in \cite{1}, \cite{2} and \cite{3}, however, in the remainder of the paper we will align with the terminology from \cite{moreno2012unifying}. Assume a model $M_i$ is trained to fit a dataset $D_i = \{ d_0, d_2, d_3, ..., d_j\}$, where $d_j = \{X_k, y_k\}$. In this sense, $\{ d_0, d_2, d_3, ..., d_j\}$ represent the data points of the dataset $D_i$, $X_k$ represents the feature vector and $ y_k$ represents the label for the corresponding data point $d_j$. Evidently, since $D_i$ can be described under a distribution $F_{0,j} (X,y)$,  $M_i$ learns to identify this distribution through the model training process. Drift can occur when new data points are inserted in $D_i$, namely $d_{j+1}, d_{j+2}, d_{j+3},.... , d_{j+n}$, if $F_{0,j} (X,y) \neq  F_{j+1,\infty} (X,y)$. For this inequality to hold true, there should be a $j$ that satisfies the following inequality: $P_j(X,y) \neq P_{j+1} (X,y)$. Since $P_j(X,y) = P_j(X) \times P_j(y|X)$, we can rewrite the drift equation as follows:

\begin{equation}
\label{eq:distr_uneq}
    \exists j: P_j(X) \times P_j(y|X) \neq P_{j+1}(X) \times P_{j+1}(y|X)
\end{equation}

Following Eq. \ref{eq:distr_uneq}, and in line with \cite{moreno2012unifying}, the four types of drifts are as following:
\begin{enumerate}

    \item The covariate shift \cite{moreno2012unifying} or drift \cite{3}, also referred to as source 1 concept drift in \cite{1}, feature drift in \cite{ackerman2022detectiondatadriftoutliers}, is observed when $P_j(X) \neq P_{j+1}(X)$, while $P_j(y|X) = P_{j+1}(y|X)$. In such cases, the feature distribution changes, when new data $\{ d_{j+1}\}$ are entered into the $D_i$ dataset, thus the reason we refer to \textit{covariate drift} also as \textit{feature drift} in this paper. 
    
    \item The prior probability shift \cite{moreno2012unifying} or drift \cite{3}  phenomenon can be identified when $P_j(x|Y) = P_{j+1}(X|y)$, while $P_j(X) \neq P_{j+1}(X)$. In this case, the label distribution changes, while in the case of the covariate drift, the distribution of the features changed.

    \item The concept shift \cite{moreno2012unifying}, drift \cite{3} or source 2  concept drift \cite{1} phenomenon can be identified when $P_j(y|X) \neq P_{j+1}(y|X)$, while $P_j(X) = P_{j+1}(X)$. In this case, the relationship between the labels and features changes. In \cite{moreno2012unifying}, it is additionally also defined as  when $P_j(X|y) \neq P_{j+1}(X|y)$, while $P_j(y) = P_{j+1}(y)$. 
    
    \item Dataset shift \cite{moreno2012unifying} or source 3 concept drift \cite{1} is  combination of covariate drift and concept drift and occurs when $P_j(y|X) \neq P_{j+1}(y|X)$ and $P_j(X) \neq P_{j+1}(X)$. This phenomenon requires both the data distribution and the feature-data mapping to change.    
\end{enumerate}

\section{Problem Statement}
\label{sec:statement}

In this paper, we focus on feature drift as defined in Section \ref{sec:drift} and assume a model $M_0$ is trained on a dataset $D_0$, which we call the \textit{original training dataset}. Given a \textit{new dataset} $D_1$, our goal is to: \textbf{(i)} asses the existence of the feature drift phenomenon; \textbf{(ii)} estimate its effects on the $M_0$ model performance and \textbf{(iii)} decide whether the $M_0$ model should be re-trained with the $D_1$ dataset in order to increase its quality.

For $M_0$, we consider two different datasets that correspond to the two distinct validation scenarios we employ in this work. Both validation scenarios leverage supervised classification tasks, one performed on a multivariate dataset (named \textit{fingerprinting}) collected from the LOG-a-TEC testbed \cite{14} and one implemented over a univariate wireless dataset (named \textit{links}) collected from the Rutgers WinLab testbed with synthetically injected anomalies/faults~\cite{9715175}. 

\begin{figure}[t]
    \centering
    \includegraphics[width=0.4\textwidth]{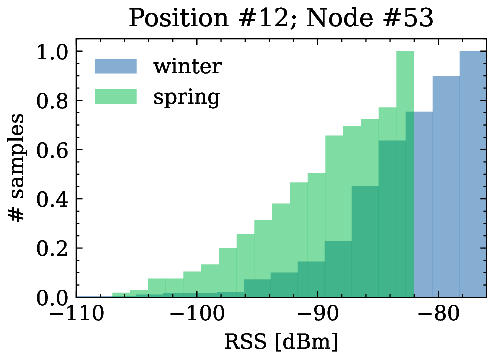}
    \caption {An example of feature drift between winter and spring data in the \textit{fingerprinting} dataset.}
    \label{fig:sprint-winter-distribution}
\end{figure}

The labels of the \textit{fingerprinting} dataset consist of discrete measurement positions in a grid and represent the location of the BLE transmitter. The dataset was collected using the LOG-a-TEC testbed in two different seasons: winter and spring. It comprises of Received Signal Strenght (RSS) data from 25 BLE nodes deployed outdoors in a campus park, with nodes mounted on light poles and building walls at varied heights. In the experiment, a BLE transmitter broadcasted signals every 100 ms across a localization grid with each grid point sampled for about one minute. The data was gathered in a realistic environment with natural ambient interference. Figure~\ref{fig:sprint-winter-distribution} represents the distribution of collected RSS measurements at node \#53 in localization position \#12 for both winter (blue bars) and spring (green bars) data. Although the two histograms partially overlap, it can be seen that the winter data range is between -102 and -78 dBm, while the range for the spring data is between -108 and -82 dBm. This shift, or feature drift, illustrates how environmental factors can alter signal propagation between seasons  even when measurements are taken at the same location and from the same transceiver pair. In this example, the testbed area is abundant with trees, bushes, and other vegetation that is fully leafed in the spring and mostly bare in the winter. The difference in foliage between seasons leads to variations in signal propagation, as the dense vegetation in spring can cause additional attenuation of the signals compared to the winter, while the absence of leaves results in less signal interference.

\begin{figure}[t]
    \centering
    \includegraphics[width=0.4\textwidth]{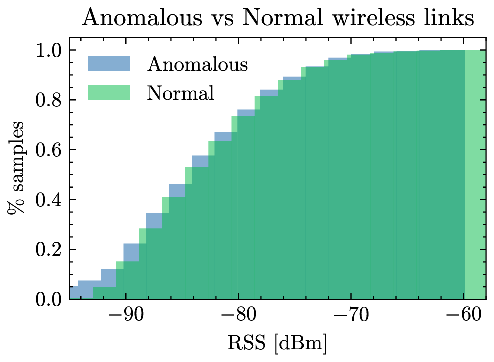}
    \caption {An example of feature drift between Normal and Anomalous wireless links in the \textit{links} dataset.}
    \label{fig:TS-normalvsanomalous-distribution}
\end{figure}

The \textit{links} dataset is a univariate wireless dataset with synthetically injected anomalies. As presented by authors in~\cite{9715175}, there are 4 common types of anomalies that can be observed in wireless link layer monitoring. As mentioned by the authors, these anomalies are rare events that can indicate different causes, such as a broken wireless nodes, software issues, or a slowly dying nodes. Figure~\ref{fig:TS-normalvsanomalous-distribution} depicts the distribution of RSS values for Anomalous (blue bars) and Normal (green bars) wireless links. As it can be seen from the figure, there is a significant overlap between both type of links, with really subtle difference between the two. The Anomalous values range between -95 to -60 dBm, while Normal values range between -92 and -58 dBm. 

We consider that the $M_0$ model is trained on the $D_0$ dataset (which can be either \textit{fingerprinting} or \textit{links}), and then it is deployed in a production environment where new data points ($D_1$) correspond to the location or type of anomaly respectively. As feature drift appears, through $D_1$, the $M_0$ responses decrease in quality. As we run a controlled experiment in which we also have labels for $D_1$, we can measure the actual decrease in performance. However, in a real production set-up, $D_1$ is a dataset which is yet to be labeled, therefore we have to develop a way to detect the feature drift in a reliable way without relying on labels.

As no labels for $D_1$ are available in production setting, detecting changes between  $D_0$ and  $D_1$ using techniques such as discussed in Section \ref{sec:drift_tech} to measure distribution changes (i.e., perform statistical tests) seems the most suitable approach. Then, we can verify which technique is the most suitable for the considered use cases. 

\section{Feature drift detection using the ALERT method}
\label{sec:data_drift_method}

\begin{figure}[t]
    \centering
    \includegraphics[width=0.4\textwidth]{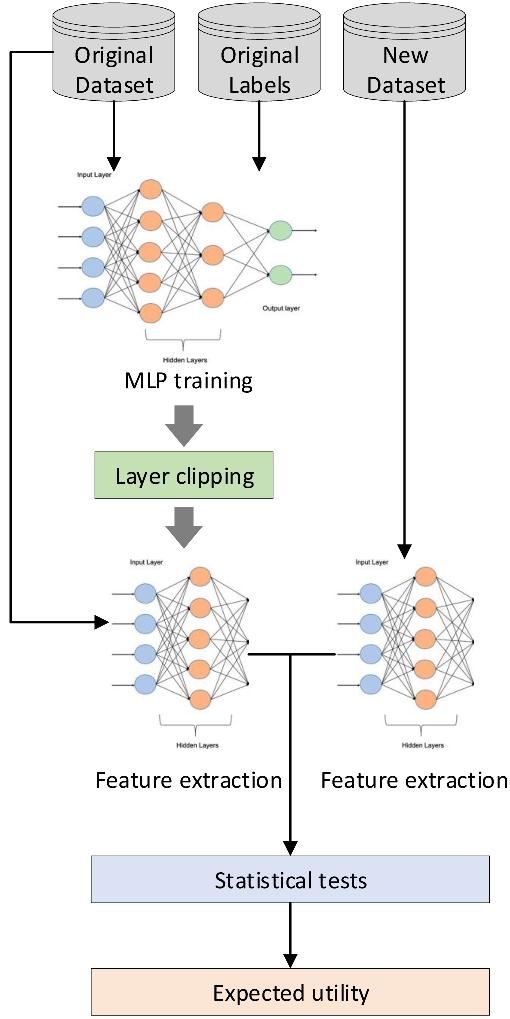}
    \caption {The proposed ALERT methodology for assessing the feature drift given 2 datasets $D_0$ and $D_1$.}
    \label{data_drift_approach}
\end{figure}

In this work we propose ALERT, a new feature drift detection method that rather than monitoring the distribution shift of the raw data or traditionally engineered features, it monitors the shift of a learnt representation (or embedding). The intuition behind ALERT is that the learnt representations tend to be lower dimensional and contain less noise making the subsequent distribution change computation faster and more accurate. ALERT includes three components: representation learning, statistical testing and utility assessment. Figure \ref{data_drift_approach} depicts the proposed method detailing the representation learning component. 

\subsection{Design of the Representation Learning Component} 

We employ a supervised approach to learn the representation of $D_0$ through a lightweight Multi-Layer Perceptron (\gls{mlp}) rather than relying on more complex and computationally expensive and data-hungry \gls{cnn} or transformer-based architectures. 

\paragraph*{Supervised representation learning} The representation is learnt by training an \gls{mlp} on $D_0$, the original dataset including the labels as depicted on the top left of Figure \ref{data_drift_approach}. Essentially the \gls{mlp} is a network of feed-forward layers where each layer consists of a number of neurons. The output $y_i$ of the $i$-th neuron can described by the following equation:

\begin{equation}
\label{eq:simple_MLP}
    y_i = \phi_i \bigg( \sum{\big( W_i X_i \big)}+ B_i \bigg)
\end{equation}

which is a linear combination of its inputs $X_i$, weights $W_i$ and biases $B_i$ where  $\phi_i $ is the activation function of the neuron. Then each layer concatenates the outputs of all its neurons $y_i$ into a vector $Y = {y_1, y_2 , ..., y_n}$, which is forwarded to the next layer. 

\paragraph*{Extracting the feature representation} For this process, we utilize the trained MLP and we clip its lower layers as depicted in the mid area of Figure \ref{data_drift_approach}. We opt to discard the lower layers of the MLP, since the upper layers tend to capture higher-level features \cite{5} and thus, they can provide useful information related with the input data distribution. In the sequel, we perform two forward passes using the clipped MLP model: one using the $D_0$ and one using the $D_1$ dataset. These forward passes do not update the weights of the model and they extract two sets of features: the $R_0$ feature set which is extracted from the $D_0$, and the $R_1$ feature set which is extracted from the $D_1$ dataset.  The extracted feature representations contain useful information that can be leveraged by the utility function designed in the next module.

 \subsection{Design of the Statistical Testing  Component} 
\label{sec:stattest}
 
For the statistical testing depicted in the blue box on the lower side of Figure \ref{data_drift_approach}, we opt to use $2$ well-established methods (\gls{kst} \cite{6,7} and \gls{psi}) \cite{8}, instead of relying on a single one. This decision aims to: \textbf{(i)} reduce the input parameters of ALERT, as much as possible; \textbf{(ii)} cancel out the weakness of \gls{kst} and \gls{psi} by aggregating their outcomes; \textbf{(iii)} reinforce ALERT with the capacity of detecting both small and medium distribution drifts; \textbf{(iv)} introduce sampling symmetry to ALERT; and \textbf{(v)} stabilize and simplify ALERT's outputs so that to be easily usable and interpretable by decision-making mechanisms.

Rather than relying on pre-set input parameters or prior knowledge regarding the (\gls{pdf}s), \gls{kst} is able to internally estimate the distribution characteristics of the input data. As a result, \gls{kst} swiftly detects small distribution  changes (even in large datasets) and produces a p-value that can be easily interpreted by external systems \cite{ks-strengths}. However, \gls{kst} is prone to false positives and is asymmetric thus, if the user swaps the baseline and sample distributions it will produce different outcomes. Also, it requires a large number of samples to accurately function and may struggle with non-normal distributions \cite{ks-limits}. \gls{psi}, on the other hand, can perform well, even if a lower amount of data is used \cite{psi-strengths-1} and is considered a stable and robust measure to assess drifts even when non-normal distributions are considered. Also, \gls{psi} produces interpretable outputs, is very good at identifying moderate or larger drifts, and is fully symmetric. On the flip side, it requires the data binning parameter to be set externally, which separates the input data into predefined intervals. The combination of \gls{kst} and \gls{psi} provides a unified framework that cancels out the weaknesses of each method and increases the robustness of ALERT. The importance of both \gls{kst} and \gls{psi}  is further discussed in Section \ref{sec:results-ablation}, in which we perform an ablation study and we showcase the individual contributions of \gls{kst} and \gls{psi} to the overall ALERT's utility score. There, it is evident that if the \gls{kst} or the \gls{psi} component is removed, the performance of ALERT drops by a large margin. To accomplish the unification of \gls{kst} and \gls{psi}, ALERT computes the following statistical tests:
\begin{itemize}
    \item The $KS_{(D_0, D_1)}$ that represents the \gls{kst} between the dataset $D_0$ and the dataset $D1$.
    \item The $KS_{(R_0, R_1)}$ that represents the \gls{kst} between the extracted features $R_0$ and the extracted features $R_1$.
    \item The $PSI_{(D_0, D_1)}$ that calculates the \gls{psi} between the dataset $D_0$ and the dataset $D_1$.
    \item The $PSI_{(R_0, R_1)}$ that calculates the \gls{psi} between the extracted features $R_0$ and the extracted features $YR_1$.
\end{itemize}

\paragraph*{\gls{kst} Statistical Test} The \gls{kst} is invoked to check if two sets of samples belong to the same distribution. To assess this, the test utilizes a $p$-value which designates that the samples belong to different distributions if $p < 0.05$. We calculate this p-value similarly with \cite{6}, as follows:

\begin{equation}
\label{eq:ks-1}
    p_{x,y} = 2 \sum_{i=1}^{z}{\big( (-1)^{i-1} \cdot e^{-2c^2(a) \cdot i^2} \big)}
\end{equation}

where $z$ is the total number of samples, $x$ and $y$ are the corresponding sets that are being checked, and the $c(a)$ can be calculated through the following formula:

\begin{equation}
\label{eq:ks-2}
    c(a) = D_{x,y} \sqrt{\frac{n_x \cdot m_y}{n_x+m_y}}
\end{equation}

where $n_x$ and $m_y$ is the number of samples of the $x$ and $y$ datasets correspondingly. $D_{x,y}$ is calculated via the following equation:

\begin{equation}
\label{eq:ks-3}
    D_{x,y} = \sup_t |F_x(t) - F_y(t)|
\end{equation}

Where $F_x(t)$ and $F_y(t)$ are the empirical distributions of the data belonging in the $x$ and $y$ sets.

\paragraph*{\gls{psi} Statistical Test} The PSI is used to measure the relative entropy between two distributions. This can be interpreted as the measurement of divergence between two different sets of samples. PSI values that are lower than $0.1$ indicate that there is no significant difference between two data distributions. We calculate PSI as suggested by previous work in \cite{8}:

\begin{equation}
\label{eq:psi}
    PSI_{x,y} =\sum_{i=1}^{z} \bigg( P(x_i)- P(y_i)  \cdot \ln \bigg( \frac{P(x_i)}{P(y_i)}\bigg)  \bigg)
\end{equation}

where $z$ is the number of samples of the $x$ and $y$ data sets, while $P(x_i)$ and $P(y_i)$ represent the frequencies of samples $i$ in the $x$ and $y$ datasets.

 \subsection{Design of the Utility Assessment  Component} 
\label{utility_explained}

Since our ultimate goal is to provide a decision-making mechanism for when to retrain the $M_0$ model, we formulate a utility function, depicted in the orange box at the bottom of Figure \ref{data_drift_approach}, that combines a \gls{kst} utility with a \gls{psi} utility as follows:

\begin{equation}
\label{eq:utility}
    U = \frac {{U_{KS}} + {U_{PSI}}}{2}
\end{equation}

 $U \in (0,1)$ encapsulates the final utility we obtain if the model retraining action is selected, given the datasets $D_0$ and $D_1$. The expected utility $U$ considers the outputs of KS and KL tests to evaluate the statistical difference of the data $D_0$ and $D_1$ and the extracted features $R_0$ and $R_1$.

Eq. \ref{eq:utility} combines Eqs. \ref{eq:util_ks} and \ref{eq:util_psi} defined as follows into a unified utility function. The KS-based utility for retraining the $M_0$ model is:

\begin{equation}
\label{eq:util_ks}
    U_{KS} = \frac{1 - KS_{(D_0, D_1)} + 1 - KS_{(R_0, R_1)}} {2} 
\end{equation}

Since $KS \in (0,1)$, $U_{KS}$ is also a bounded function ($U_{KS} \in (0,1)$). This function uses the information derived from the datasets $D_0$ and $D_1$ and averages it with the information extracted from the features $R_0$ and $R_1$ to assess the KS drift. Evidently, when $U_{KS} \to 1$ the drift phenomenon is more prominent, while when $U_{KS} \to 0$ no drift is detected. 

We also devise a function to calculate the PSI-based utility for retraining the $M_0$ model:

\begin{equation}
\label{eq:util_psi}
    U_{PSI} = \sigma \big ( \frac{PSI_{(D_0, D_1)} + PSI_{(R_0, R_1)}}{2} \big)
\end{equation}

The $U_{PSI}$ averages the $PSI_{(D_0, D_1)}$ and the $PSI_{(R_0, R_1)}$ and uses the sigmoid function $\sigma$ to bound the result so that $U_{PSI} \in (0,1)$. Similarly to Eq. \ref{eq:util_ks}, the expected utility of the model retrain operation is higher when $U_{PSI} \to 1$ and lower when $U_{PSI} \to 0$.

\section{Evaluation Methodology}
\label{sec:eval}

\subsection{Dataset Description}
For our experiments we  focus on two use cases with two  datasets: \textit{fingerprinting} dataset summarized in Figure \ref{fig:sprint-winter-distribution}, and the \textit{links} dataset summarized in Figure \ref{fig:TS-normalvsanomalous-distribution}. 

\begin{table}[t]
\caption{The details of the drifted and non-drifted datasets which are created based on the \textit{fingerprinting} and \textit{links}.}
\label{tab:dataset_creation}
\resizebox{.49\textwidth}{!}{
\begin{centering}
  \begin{tabular}{c|cc}
& \textit{fingerprinting} & \textit{links}\\
\hline
Data types & BLE measurements & Time series\\
Total samples & $505.000$ & $8.493 \times 302$\\
\# Classes & $25$ & $5$\\
$M_0$ training & $D_0$ & $D_0$\\
\hline
 & $D_3, D_6, D_9, D_{12},$ & $D_4, D_5, D_6,$\\
Drifted data& $ D_{15}, D_{18}, D_{21}, D_{24},$ & $D_7, D_8$ \\
& $ D_{27}, D_{30}$ &  \\
\hline
& $D_1, D_2, D_4, D_5, $ & \\
& $D_7, D_8, D_{10}, D_{11},$ & \\
Non-drifted data& $D_{13}, D_{14}, D_{16}, D_{17},$ & $D_1, D_2, D_3$ \\
& $D_{19}, D_{20}, D_{22}, D_{23},$ & \\
& $D_{25}, D_{26}, D_{28}, D_{29}$ & \\
\hline
Cause of drift & Spring data is & New anomalies are\\
& mixed with winter data & added to the data\\
\end{tabular}
\end{centering}
}
\end{table}

The \textit{fingerprinting} dataset \cite{14} contains received signal strength (RSS) measurements made with Bluetooth Low Energy (BLE) technology, measured in dBm. The dataset consists of $505.000$ data points, organised over $25$ classes that represent 2D coordinates, which are collected during the spring and during the winter, as described in Section \ref{sec:statement}. It can be used for outdoor fingerprint-based localization applications, similarly to \cite{14}. We organise the data into $31$ smaller datasets each  containing $16.290$ samples, as follows: The first dataset $D_0$, which contains samples collected during the winter season only, is used to train a random forest classifier, as our $M_0$ model. For this reason, we refer to the $D_0$ as the \textit{original dataset}, as depicted in Figure \ref{data_drift_approach}. Then we edit the rest of the datasets $D_1$ - $D{30}$ so that each third dataset contains samples stemming from spring measurements, which are drifted (i.e., $D_3$, $D_4$, $D_6$, $D_7$, $D_9$ etc.). The rest of the datasets (i.e., $D_2$, $D_5$, $D_8$, $D_{11}$, $D_{14}$ etc.) contain samples that are collected during winter and belong to the same distribution with $D_0$.  Through this process, we can simulate various scenarios of feature drifts which are encountered in different time frames i.e., in different versions of datasets collected in the field. Table \ref{tab:dataset_creation} summarizes the drift creation details for the \textit{fingerprinting} dataset.

\begin{figure}[t]
    \centering
    \includegraphics[width=0.4\textwidth]{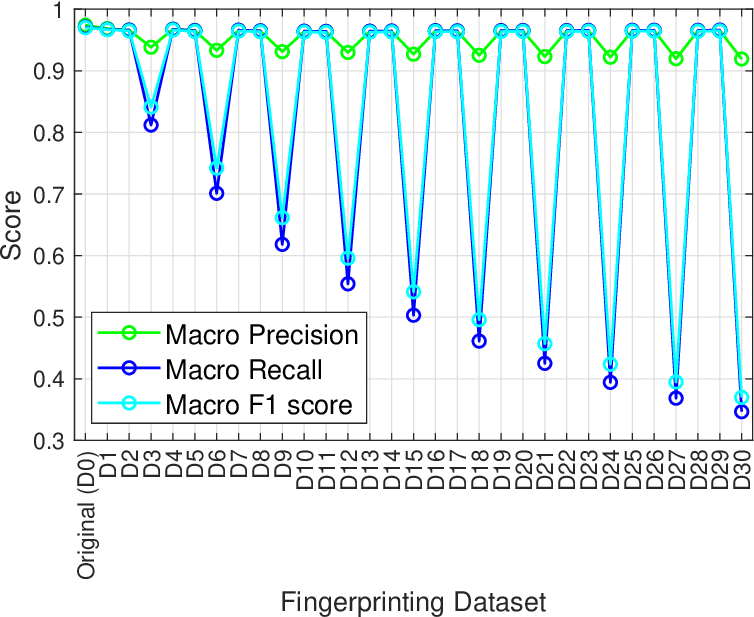}
    \caption { The performance of the $M_0$ model with different \textit{fingerprinting} datasets. The performance drops when the feature drift phenomenon is present, since the $M_0$ is trained with the $D_0$ dataset.}
    \label{fig:sprint-winter-baseline}
\end{figure}

In the sequel, we test the $M_0$ model's Macro Precision, Macro Recall and Macro F1-score with each created dataset ($D_1 - D_{30}$). As illustrated in Figure \ref{fig:sprint-winter-baseline}, the performance of the $M_0$ drops significantly when it is tested with a dataset that contains drifted samples. This experimental set-up shows the existence of the drift and enables its detection through ALERT and selected baselines.
\begin{figure}[t]
    \centering
    \includegraphics[width=0.4\textwidth]{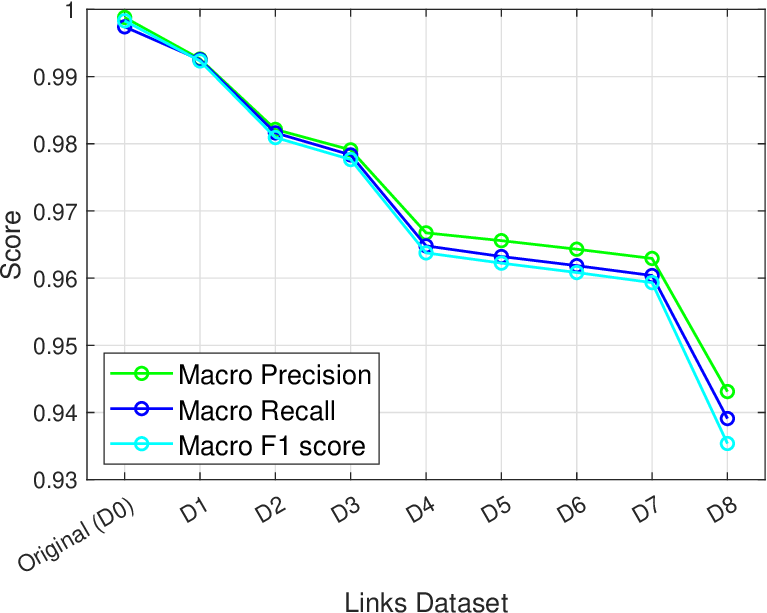}
    \caption {The performance of the $M_0$ model with different \textit{links} datasets. The model performance deteriorates when the $M_0$ is tested with datasets that contain feature drifts.}
    \label{fig:ts-baseline}
\end{figure}

The \textit{links} dataset contains data from $8492$ timeseries, each one of which having $302$ data samples. The dataset is labeled and is organized into $5$ anomaly classes. We split the data into $9$ smaller datasets each one containing $943$ timeseries, as follows: the first dataset $D_0$ is used to train a random forest classifier, as our $M_0$ model. Similarly to our approach when using the \textit{fingerprinting} case, the $D_0$ is again the \textit{original dataset}, which is illustrated in Figure \ref{data_drift_approach}. Then we split the rest of the datasets $D_1$ - $D_8$ so that the $D_1$, $D_2$ and $D_3$ to contain samples from the same distribution as $D_0$, while the $D_4$, $D_5$, $D_6$, $D_7$ and $D_8$ to contain drifted samples, from new anomaly types. Table \ref{tab:dataset_creation} summarizes the details of this process. We then test the $M_0$ model's Macro Precision, Macro Recall and Macro F1-score with each created dataset and we present our findings in Figure \ref{fig:ts-baseline}. We observe that the performance of the $M_0$ drops when used with drifted samples, similarly to the \textit{fingerprinting} model. 

\subsection{Parameter Search for the Representation Learning Components}
In order to design an efficient model for representation learning, we should consider tuning the  \gls{mlp}'s layers, number of neurons contained in each layer and the number of training epochs. It is our objective to avoid complicated models for the representation learning process, due to the computational and training requirements they would impose during the \gls{mlp} training operation. For this reason, we try to minimize as much as possible the number of layers, the amount of neurons and the training epochs of the model. On the contrary, we are aware that we risk underfitting if we design the model to be very simple since, in that case it would be unable to capture the representations of the input data. To solve this issue, we perform a $3$-dimensional parameter search regarding the number of layers, neurons and training epochs.

\subsection{Baseline Selection}
\label{sec:eval-baselines}
We compare the ALERT technique with state-of-the art methods that exist in the literature. We choose three broader types of methods for comparison, namely statistics-based methods, distance-based methods and ML-based methods. Statistical methods leverage statistical indexes (such as mean values, sampling variations and variance) to predict feature drifts. In this work, we formulate a baseline using the following statistical methods: (i) Kuiper test \cite{kuiper-ref}; (ii) Cramer-Von Mises (CVM) \cite{cvm-ref}; (iii) Welch Test (WelchT) \cite{welch-ref}; (iv) Chi Square test \cite{chi-ref}; (v) Mann-Whitney U test (Mann Whitney) \cite{man-ref}; (vi) Andreson Darling Test \cite{and-ref}; and (vii) Kolmogorov-Smirnov (KS) test \cite{ks-ref}. On the contrary, distance-based methods focus on estimating the distance between two data distributions by measuring  the dissimilarity between them through distance functions. In this work we use as a baseline the following distance-based methods: (i) Population Stability Index (PSI) \cite{psi-ref}; (ii) Energy Distance \cite{energy-distance-ref}; and (iii) Earth Mover's Distance (EMD) \cite{emd-ref}. ML-based methods utilize Deep Learning or Machine Learning models to either form an intermediate representation of the data under investigation, or to estimate the features of data distributions. In this paper we use the following ML-based methods as baselines: (i) Adversarial Detection (ML-AD) \cite{adversarial_baseline}; (ii) Embedding-based domain classification (ML-EDC) \cite{embedding_baseline}; and (iii) Data-driven deep Density Estimation (ML-DDE) \cite{density_baseline}.

\subsection{Training and Evaluation}

We use the $D_1$ - $D_{30}$ stemming from the \textit{fingerprinting} and the $D_1$ - $D_8$ stemming from the \textit{links} datasets to evaluate ALERT and to assess whether it is able to identify the feature drift phenomenon accurately. 
For the implementation of ALERT, we use the python programming language; whereas the code base for the Baseline methods is provided by \cite{adversarial_baseline}, \cite{embedding_baseline}, \cite{density_baseline} and \cite{CESPEDESSISNIEGA2024101733}. For each validation Use Case, we train the ALERT method's \gls{mlp} for $3$ epochs using the $D_0$ dataset and then, we utilize the trained model to assess the existence of the feature drift phenomenon.

Since the distance-based and statistics-based methods are provided in a ready-to-use form in GitHub repositories \cite{CESPEDESSISNIEGA2024101733}, their evaluation is trivial. On the other hand, ML-based methods require some adaptations in order to properly work with our datasets. For this reason, we have performed the following adaptations:   

\paragraph*{For the ML-AD method}  we use the open source code provided by the author \cite{adversarial_baseline} and we opt to train a ResNet-28-10 model from scratch to fit the $D_0$ dataset. Then, we fine-tune the model using the provided Bayesian ensemble method, which utilizes one optimizer (SGD) and one regularizer (Prior Regularizor) in parallel. In the sequel, we save the trained model and test its performance with each new dataset ($D_1$ - $D_{30}$ for the \textit{fingerprinting} data, or $D_1$ - $D_8$ for the \textit{links} data). This approach essentially performs adversarial detection for each new dataset. The outcomes of each test determine if the drift phenomenon is detected. Thus, when the model detects adversarial samples, we assess that the drift phenomenon is present.

\paragraph*{For the ML-EDC method} we use an unsupervised representation learning approach for drift detection. More specifically, we train the DriftLens model~\cite{embedding_baseline} with our $D_0$ dataset in order for the model to learn the distribution properties of the input data. Then, we utilise the $D_1$ dataset to estimate the threshold distance values that discriminate between drift and no-drift conditions, as suggested by the authors. Afterwards, we utilise the rest of the datasets to perform inference of the trained model.
    
\paragraph*{For the ML-DDE method} we first train a new model to recognize the Probability Density Function (PDF) characteristics of our $D_0$ dataset. For the \textit{fingerprinting} dataset we use a $2$-Dimensional data structure ($X$ and $Y$ coordinates), while for the \textit{links} dataset we use a $5$-Dimensional data structure ($1$ dimension for each class). In the sequel we conduct inference using the trained model and the rest of the datasets. In order to assess the presence of drift phenomena, we use the KL-divergence metric, as suggested by the authors \cite{density_baseline}, to measure the distance between the estimated PDF of the $D_0$ and the estimated PDF of the dataset under investigation.

\subsection{Performance Metrics}
\label{sec:metrics}
Since ALERT uses a utility function that designates whether the $M_0$ model should be retrained or not, it is difficult to compare it with existing solutions. This happens because each state-of-the-art method uses a different prediction confidence threshold that is uniquely tailored according to its requirements. To resolve this, we define the following scoring function that can be commonly used among several methods to compare their efficiency:

\begin{equation}
\label{eq:performance-metric}
    Score=\begin{cases}
    F1_{gain},      & \text{if decision is true positive}.\\
    T_s,      & \text{if decision is true negative}.\\
    F1_{gain}-T_s,  & \text{if decision is false positive}.\\
    -F1_{gain},  & \text{if decision is false negative}.\\
    \end{cases}
\end{equation}

The function is designed to increase the score for correct feature drift predictions and to decrease for incorrect ones. The larger the feature drift, the bigger score is allocated for correct assessments, and the bigger the penalty is given for incorrect predictions. We assume that each method under examination outputs a prediction on whether the $M_0$ should be retrained. We distinguish $4$ different possible scenarios for this assessment:

\begin{itemize}
    \item A method's assessment is true positive and correctly identifies the existence of feature  drift. There, the score  equals to the macro $F1$-score that the model will gain if retrained ($F1_{gain}$). The larger the feature  drift, the higher the ($F1_{gain}$), and thus, larger scores  are allocated for correct assessments.
    \item A method's assessment is true negative and correctly identifies the absence of feature drift. In such scenarios the score equals to the $T_s$ constant, that is set by the user. In our experiments, we set the $T_s$ to $0.1$. 
    \item A method's assessment is false positive and incorrectly identifies the existence of feature drift. In this case the model $M_0$ is retrained, and the method is penalized by an amount of $F1_{gain}-T_s$ where $F1_{gain}$ is $F1$-score the model gains after retraining. We expect a small $F1_{gain}$, due to the absence of feature  drift and as a result, the score will be negative. 
    \item A method's assessment is false negative and incorrectly identifies the absence of feature  drift. The penalty of this error is $-F1_{gain}$, which is the $F1$-score that the model would gain if it was retrained with the new data. 
\end{itemize} 

In our experiments, we apply this formula for each method under examination and for each tested dataset. At the end of each experiment, we sum each method's scores which are collected after assessing the aforementioned datasets and we calculate the final value.  

\subsection{Utility Component Contribution}
\label{sec:eval-ablation}
We perform an analysis of the contribution of each of the following components to the total utility score of the ALERT method: (i) $KS_{(D_0, D_1)}$; (ii) $KS_{(R_0, R_1)}$; (iii) $PSI_{(D_0, D_1)}$; and (iv) $PSI_{(R_0, R_1)}$. The aim of this analysis is to validate our initial hypothesis that both the \gls{kst} and the \gls{psi} tests are essential for the calculation of the utility score. Through this, we also aim to analyze the impact of the representation learning technique to the overall utility score of the ALERT. To achieve this, we perform an ablation study by measuring the contributions of each component separately, in terms of percentages ($\%$) for each Use Case, and we present the results we obtain in Section \ref{sec:results-ablation}.

\section{Results}
\label{sec:results}
In this section we analyze the performance of the ALERT method proposed in Section \ref{sec:data_drift_method} and evaluated according to the methodology elaborated in Section \ref{sec:eval} to solve the problem identified in Section \ref{sec:statement}.

 \begin{figure}[t]
    \centering
    \includegraphics[width=0.4\textwidth]{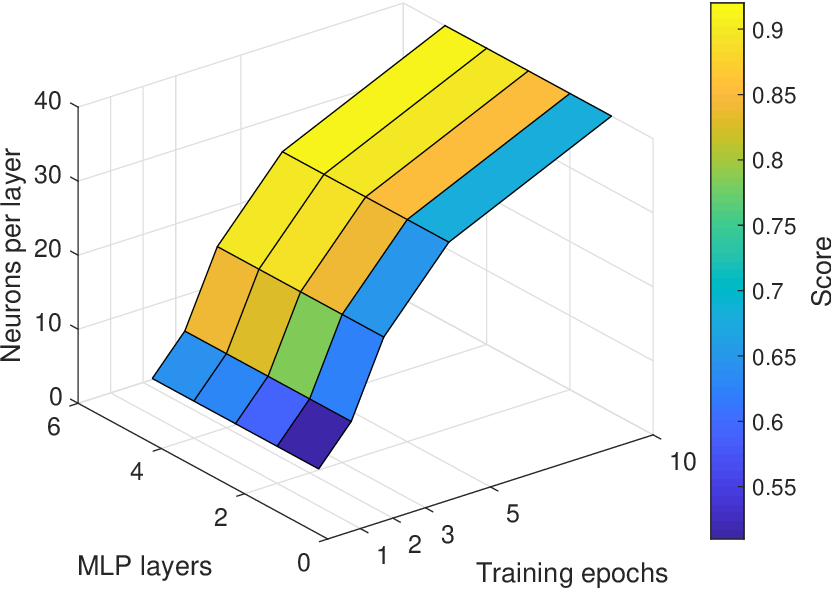}
    \caption {The contributions of different \gls{mlp} parameters and training epochs to the obtained utility score.}
    \label{mlp_parameters}
\end{figure}
\begin{table}[thb]
\caption{The MLP parameters used for the implementation of ALERT.}
\label{tab:mlp_param}
\centering
\resizebox{.75\columnwidth}{!}{
  \begin{tabular}{c|c}
MLP parameters & Values \\
\hline
Layers & 3 (Dense, Dense, Dense) \\
Neurons & 20, 20, Number of Classes \\
Activations & Sigmoid, Sigmoid, Softmax \\
Training epochs & 3\\
Batch size & 20 \\
Optimizer & Adam \\
Loss & Categorical Cross-Entropy
\end{tabular}
}
\end{table}

\subsection{Parameter Search for the Representation Learning Components}
\label{sec:results-parasearch}
Figure \ref{mlp_parameters} illustrates the results of the $3$-dimensional parameter exploration, which is conducted for the \gls{mlp}. We experiment with two datasets in which the feature drift phenomenon is prominent, in order to calibrate the \gls{mlp} parameters. We should note that the utility function should be represented by a high value, to clearly predict the existence of feature drifts. Results indicate that the increase of model complexity is directly correlated with higher utility scores. This is expected, since more complex models manage to properly capture input data representations and thus, to produce better assessments. Nonetheless, there is an optimal parameter space, over which larger and more complex \gls{mlp}s do not provide significant contributions to the utility score. Empirically, an \gls{mlp} with $3$ layers, $20$ neurons per layer, and a training period of $3$ epochs achieves a utility score of $0.85$, which we deem adequate to assess the presence of feature drifts. On the contrary, an \gls{mlp} with $5$ layers, $40$ neurons per layer, and a training period of $10$ epochs achieves a utility score of $0.92$ which, despite being higher than $0.85$, does not provide us with better information than the previous configuration. This is true especially if we consider that the training requirements of more complex \gls{mlp} are significantly larger compared to models that leverage simpler designs.

Table \ref{tab:mlp_param} lists the \gls{mlp} parameters used for the ALERT method. As discussed in the paragraph above, we opted for a small model with $3$ layers, each one of which containing $20$ neurons. Since ALERT is a supervised method, the last layer contains a number of neurons equal to the amount of data classes. We utilise the Sigmoid activation function for the first two layers and the Softmax function for the last layer correspondingly. We train the model for $3$ epochs, using a batch size of $20$, selecting Adam as the optimiser and the Categorical cross-entropy as loss. Due to the small size of the \gls{mlp} we do not employ any regularization methods, and since the training is conducted for $3$ epochs only, we do not use any early-stopping techniques. In order to avoid temporal leakage, we use the $D_0$ dataset for training, and the rest of the datasets for inference. This way, future data points are not accounted for, during the model training.
\begin{figure*}[t]
    \centering
    \includegraphics[width=0.99\textwidth]{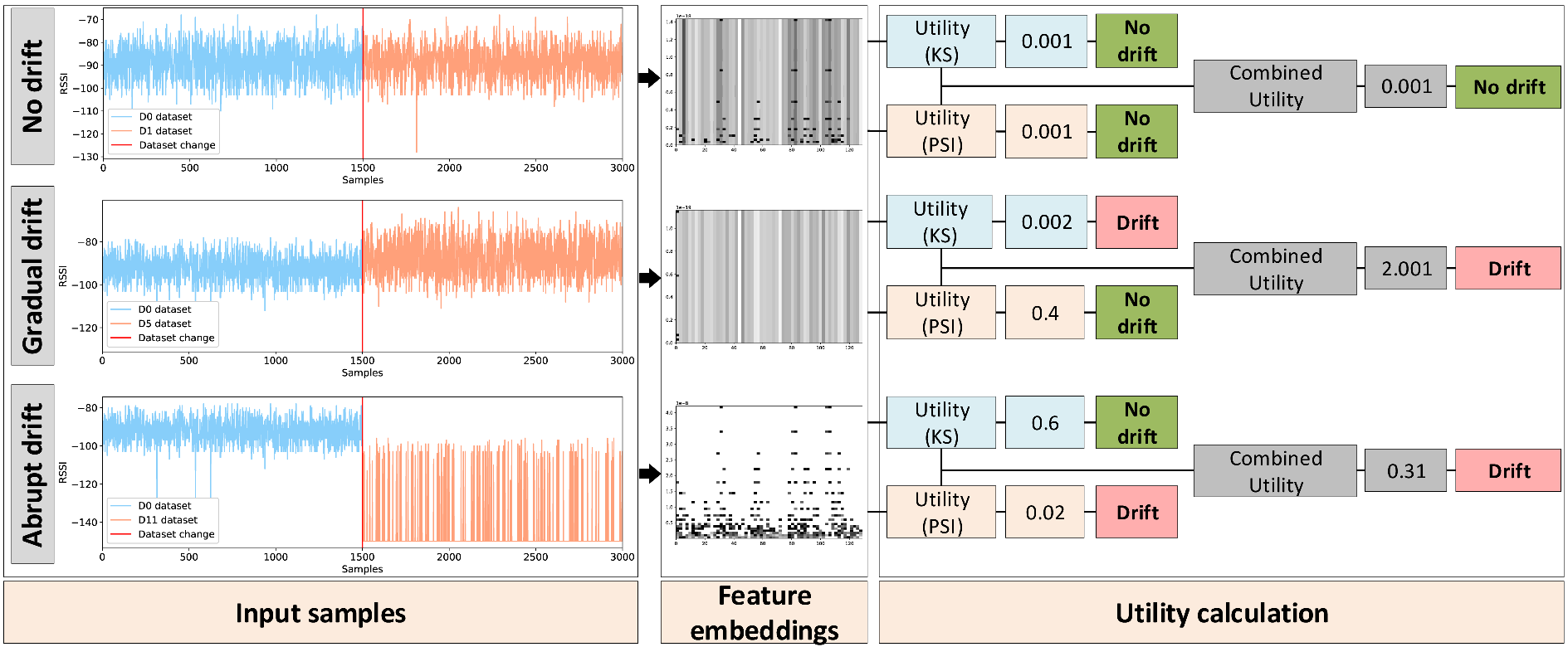}
    \caption {The behavior of \gls{psi} and \gls{kst} in different drift scenarios (no drift, gradual drift and abrupt drift). \gls{psi}  excels in identifying abrupt drifts, while \gls{kst}  is good at detecting gradual drifts. The combination of both detectors, along with the proposed representation learning technique give ALERT better coverage for multiple drift scenarios.}
    \label{fig:eg}
\end{figure*}

\subsection{Performance under different feature drift scenarios}

Figure \ref{fig:eg} illustrates the performance of ALERT under different different drift scenarios. To properly assess the effects of \gls{kst} and \gls{psi} in the design of the utility function in Eq. \ref{eq:utility} we opt to visualise their outputs for $3$ scenarios, using the \textit{fingerprinting} dataset, which is illustrated in Figure \ref{fig:sprint-winter-distribution}: (i) when no drift exists in the data; (ii) when a gradual drift occurs; and (iii) when an abrupt drift appears. Evidently, for each scenario, ALERT produces different feature embeddings which are then interpreted by the utility calculation module. As discussed in Section  \ref{utility_explained}, this module utilises both \gls{kst} and \gls{psi} to assess if a drift phenomenon exists. The results confirm the complementarity of \gls{kst} and \gls{psi} incorporated in Eq. \ref{fig:eg}. More specifically, the feature analysis conducted by \gls{kst} achieves good results when gradual drifts are under examination, while in such cases \gls{psi} underperforms and thus, it cannot properly identify their presence. The opposite observation for abrupt drifts holds true. There,  \gls{psi} excels in drift detection, whereas \gls{kst} often fails to identify abrupt drifts. To solve this, ALERT combines the outputs of both estimators and manages to capture both abrupt and gradual drifts. ALERT, \gls{kst} and \gls{psi}  behave similarly  when dealing with the \textit{Links} dataset as well, which is illustrated in Figure \ref{fig:TS-normalvsanomalous-distribution}.

\begin{figure}[t]
    \centering
    \includegraphics[width=0.4\textwidth]{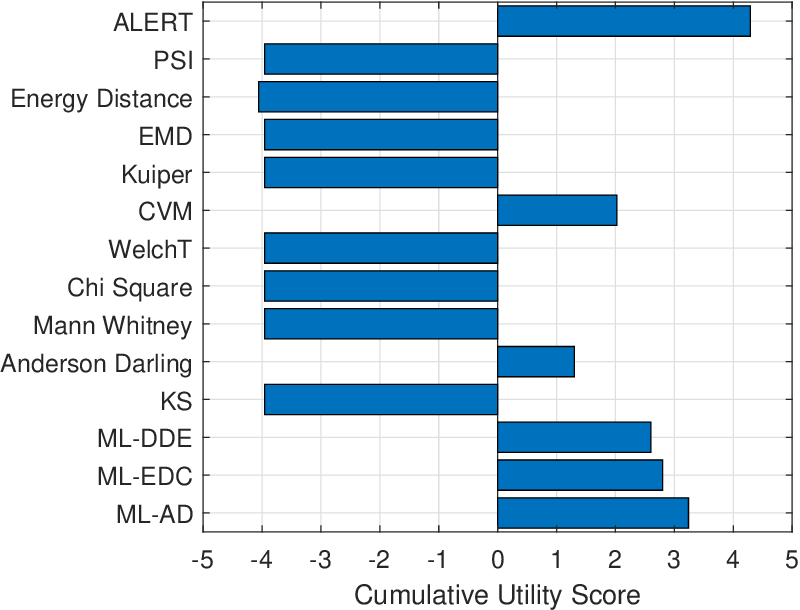}
    \caption { Performance comparison between the ALERT and the baseline tests, using the \textit{fingerprinting} dataset.}
    \label{fig:mlp-spring}
\end{figure}

\begin{table*}[tbh]
\caption{Performance in terms of Precision, FRD, Recall and F1 of different drift detection methods over the \textit{fingerprinting} dataset. The proposed ALERT method outperforms all the baseline models, achieving an overall F1-score of $0.9$. In terms of F1-scores, ML-AD follows with $0.66$, ML-DDE  with $0.64$ and ML-EDC with $0.0.64$}
\label{tab:spring-pred}
\resizebox{.99\textwidth}{!}{
\begin{centering}
  \begin{tabular}{c|cccccccc}
  Methods               & True positives       & True negatives    & False positives     & False negatives & Precision & FDR & Recall & F1\\
  \hline
  ALERT            & $9/10$    & $19/20$   & $1/20$    & $1/10$ & $0.9$   & $0.1$& $0.9$  & $0.9$\\
  CVM            & $8/10$   & $10/20$    & $10/20$    & $2/10$ & $0.44$   &$0.56$& $0.8$  & $0.56$\\
  Anderson Darling   & $5/10$    & $11/20$   & $9/20$    & $5/10$  & $0.35$  &$0.65$& $0.5$  & $0.41$\\
  PSI               & $2/10$    & $2/20$   & $18/20$    & $8/10$ & $0.1$  &$0.9$& $0.2$  & $0.13$\\
  ML-AD  & $9/10$ & $12/20$    & $8/20$    & $1/10$  & $0.53$   &$0.47$& $0.9$  & $0.66$\\
    ML-EDC           &  $8/10$ & $15/20$  & $5/20$  & $2/10$ & $0.61$  &$0.39$& $0.8$  & $0.62$ \\
  ML-DDE     & $8/10$    & $13/20$   &  $7/20$   & $2/10$ & $0.53$  &$0.47$&   $0.8$ & $0.64$\\
  \end{tabular}
  \end{centering}
}
\end{table*}
\begin{figure}[t]
    \centering
    \includegraphics[width=0.4\textwidth]{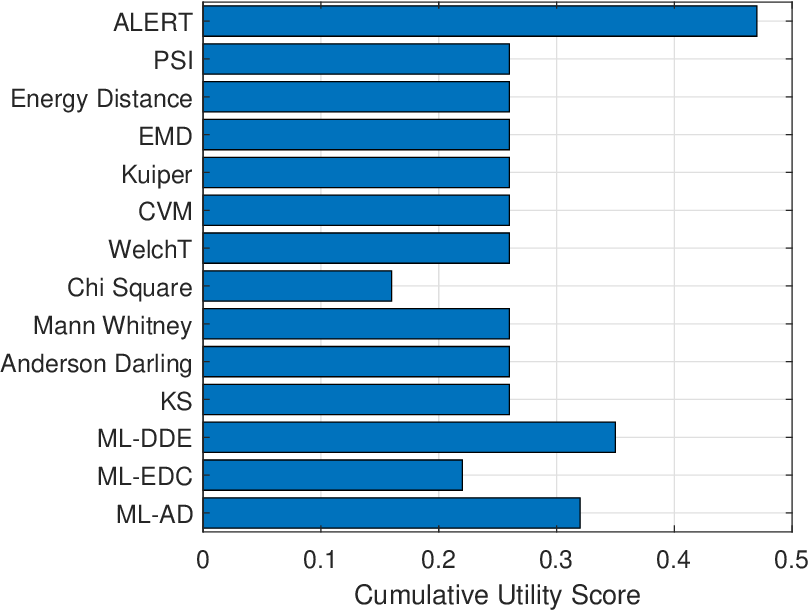}
    \caption { Performance comparison of the ALERT method, with baseline techniques, using the \textit{links} dataset.}
    \label{fig:mlp-ts}
\end{figure}
\begin{table*}[thb]
\caption{Performance in terms of Precision, FRD, Recall and F1, of different drift detection methods over the \textit{Links} dataset. ALERT achieves the best F1-score, namely $0.88$. ML-DDE achieves $0.75$, ML-AD $0.72$, CVM and PSI $0.6$}
\label{tab:ts-pred}
\resizebox{.99\textwidth}{!}{
\begin{centering}
  \begin{tabular}{c|cccccccc}
  Methods               & True positives       & True negatives    & False positives     & False negatives & Precision & FDR &Recall & F1\\
  \hline
  ALERT            & $4/5$    & $3/3$   & $0/3$    & $1/5$ &$1.0$ & $0$& $0.8$ & $0.88$ \\ 
  CVM            & $3/5$   & $1/3$    & $2/3$    & $2/5$   &$0.6$  & $0.4$& $0.6$ & $0.6$ \\ 
  Anderson Darling   & $2/5$   & $2/3$    & $1/3$    & $3/5$   &$0.66$  & $0.34$& $0.4$ & $0.49$\\
  PSI               & $3/5$   & $1/3$    & $2/3$    & $2/5$   &$0.6$  & $0.4$& $0.6$ & $0.6$\\
  ML-AD  & $4/5$ & $1/3$    & $2/3$    & $1/5$  & $0.66$   & $0.34$& $0.8$  & $0.72$\\
  ML-EDC      & $2/5$   & $3/3$  & $0/3$  & $3/5$ & $1.0$  & $0$& $0.4$   & $0.57$  \\
  ML-DDE     &  $3/5$  &  $2/3$ &   $0/3$  & $2/5$ & $1.0$  & $0$& $0.6$   & $0.75$  \\
  
  \end{tabular}
  \end{centering}
}
\end{table*}

\subsection{Performance with the "Fingerprinting" Dataset}
\label{sec:results-Fingerprinting-performance}
Figure \ref{fig:mlp-spring} depicts the performance, in terms of the score function established by the Eq.  \ref{eq:performance-metric}, for the proposed ALERT method and the baseline techniques described in Section \ref{sec:eval-baselines}. The maximum achievable score for this Use Case is $4.5$, which is achieved if all the predictions are correct, and the lowest is $-4.5$ which is obtained if all the predictions are incorrect. We observe that the proposed ALERT method achieves the best score ($4.28$), followed by  the ML-AD ($3.24$), ML-EDC ($2.8$), ML-DDE ($2.6$), CVM ($2.0$) and Anderson Darling ($1.3$) tests. The rest of the methods underperform by a large margin and they obtain negative scores. The superiority of ALERT is due to the learnt representation that keeps only the relevant signals related to drift, and to the design of the utility function in Eq. \ref{eq:utility} that provides balanced sensitivity, as discussed in Section \ref{utility_explained}.  This also contributes to ALERT's robustness, reducing false positives and false negatives in drift detection.

Table \ref{tab:spring-pred} depicts the detailed prediction statistics for the four best performing methods on the "\textit{fingerprinting"} dataset. The first column of the table represents the methods under examination, while the rest of the columns depict each method's performance in terms of true positive, true negative, false positive and false negative predictions, as described in Section \ref{sec:metrics}. We have also included $4$ additional columns that present the precision, false discovery rates (FDR), recall and F1-score of each method. The ALERT method outperforms the baseline techniques since it achieves the best F1-score, namely $0.9$. ML-based methods follow with ML-AD ($0.66$), ML-DDE ($0.64$) and ML-EDC ($0.62$). The rest of the methods perform poorly, as the CVM achieves $0.56$ F1, the Anderson Darling $0.41$ and the PSI $0.13$. Evidently, ALERT manages not only to accurately detect the existence of drift phenomena ($9/10$ true positive score), but also to correctly assess the absence of feature  drift ($19/20$ true negative score), thus saving the $M_0$ model of unnecessary re-training operations. Also, ALERT minimizes the rate under which it falsely identifies drifts, achieving a low FDR score ($0.1$). ML-AD, which is the best performing baseline method, achieves the same true positive prediction score ($9/10$), but a lower true negative score ($12/20$). Further, ML-AD has a higher FDR score ($0.47$) which would result in unnecessarily re-training operations of the $M_0$ model. The same observation also holds true for both the rest of the baseline methods.

\subsection{Performance with the "Links" Dataset}
\label{sec:results-Links-performance}
Figure \ref{fig:mlp-ts} illustrates the comparison of the ALERT technique with existing works over the \textit{links} dataset. The maximum score for each method is $0.48$. The ALERT achieves a score of $0.47$ and outperforms the rest of the baseline techniques. Distance-based and statistics-based methods acquire an equal score of $0.26$, with the exception of Chi-Square that severely under-performs ($0.16$ score). On the other hand, ML-based techniques perform well, with ML-DDE leading with $0.35$  followed by ML-AD ($0.32$) and ML-EDC ($0.22$). ALERT manages to successfully capture the data distribution properties of the \textit{links} dataset, containing both gradual and abrupt drifts, and to assess with high accuracy the existence of feature drift.

Table \ref{tab:ts-pred} showcases the performance of top-scoring methods, considering their predictions. Similarly to Table \ref{tab:spring-pred}, the first column depicts the methods under examination which are the ALERT, CVM, Anderson Darling, PSI, ML-AD, ML-EDC and ML-DDE. The rest of the table columns contain information regarding the true positive, true negative, false positive, false negative, precision, FDR, recall and F1 scores of each method. The ALERT and the ML-AD technique achieve the best true positive score ($4/5$), followed by ML-DDE, CVM and PSI ($3/5$). In terms of true negatives, ALERT and ML-EDC come first with $3/3$ correct assessments, Anderson Darling and ML-DDE second with $2/3$ and the rest follow with $1/3$. This has a direct impact to each method's F1, with ALERT scoring $0.88$, closely followed by ML-DDE ($0.75$) and ML-AD ($0.72$), while CVM and PSI score $0.6$. We also observe that all ML-EDC, ML-DDE and ALERT achieve $0$ FDR scores and thus, they minimize the unnecessary model re-training operations. Apart from the FDR, the rest of results are in line with the results we observed under the \textit{fingerprinting} dataset. The ALERT method not only identifies the existence of drift phenomena, but also it properly assesses their absence as well.

\begin{figure}[t]
    \centering
\includegraphics[width=0.4\textwidth]{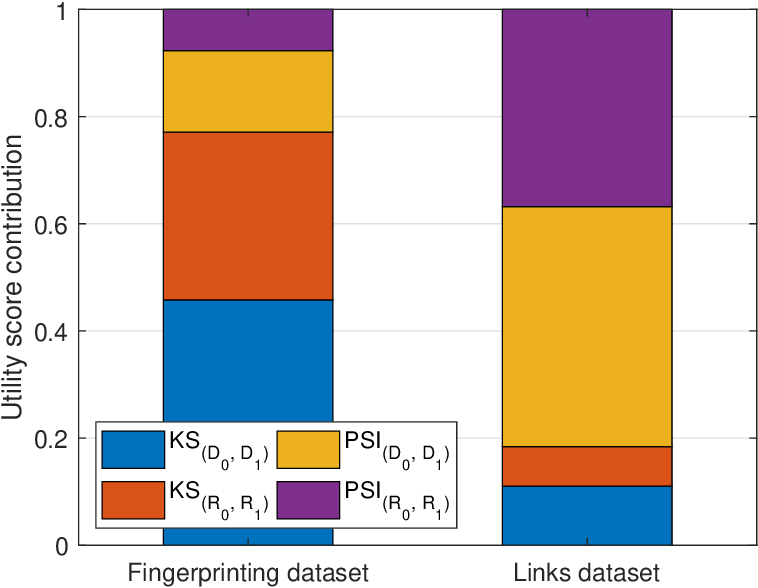}
    \caption {The contribution of $KS_{(D_0, D_1)}$, $KS_{(R_0, R_1)}$, $PSI_{(D_0, D_1)}$ and $PSI_{(R_0, R_1)}$ to the expected utility score.}
    \label{statistics_ablation}
\end{figure}
\subsection{Ablation Study for the Utility Assessment Component}
\label{sec:results-ablation}
Figure \ref{statistics_ablation} depicts the outcomes of the ablation study, which we conducted in order to evaluate the contributions of the proposed representation learning technique along with the \gls{kst} and \gls{psi} tests described in Section \ref{sec:stattest} to the utility score of ALERT. As discussed in Section \ref{sec:eval-ablation}, we perform the study using the \textit{fingerprinting} and the \textit{links} datasets to confirm our hypothesis that all of the aforementioned methods play a major role to the overall utility score. Results indicate that the contributions of each statistical test ($KS_{(D_0, D_1)}$, $KS_{(R_0, R_1)}$, $PSI_{(D_0, D_1)}$ and $PSI_{(R_0, R_1)}$) are indeed significant. Evidently, all four methods have quantifiable impact to the expected utility, ranging from $7\%$ to $45\%$ depending on the dataset. Therefore, the exclusion of any of these tests would reduce the effectiveness of the utility function for the ALERT method.

\begin{table}[h]
\caption{The execution time requirements of the methods under examination for both the \textit{fingerprinting} and \textit{links} datasets. The execution time is measured in seconds and is calculated for each method independently. 
}
\label{tab:exec-time}
\resizebox{.45\textwidth}{!}{
\begin{centering}
  \begin{tabular}{c|cccc}
  Methods               & \textit{fingerprinting} dataset       & \textit{links} dataset     \\
  \hline
  ALERT                   & $17.2s$  & $3.5s$          \\  
  PSI                   & $0.04s$  & $0.4s$          \\
  Energy Distance       & $0.04s$  & $0.3s$          \\
  EMD                   & $0.04s$  & $0.3s$          \\
  Kuiper                & $0.07s$  & $0.3s$          \\
  CVM                   & $0.03s$  & $0.19s$          \\
  WelchT                & $0.02s$  & $0.02s$          \\
  Chi Square            & $0.06s$  & $0.4s$          \\
  Mann Whitney          & $0.03s$  & $0.1s$          \\
  Anderson Darling      & $0.03s$  & $0.3s$          \\
  KS                    & $0.06s$  & $0.2s$          \\
  ML-AD            & $22$s & $4s$  \\
  ML-EDC           & $25$s & $5.2s$    \\
  ML-DDE           & $28$s & $7s$   \\
  
  \end{tabular}
  \end{centering}
}
\end{table}

\subsection{Execution time requirements}
\label{sec:results-time}
In Table \ref{tab:exec-time} we present the execution time requirements of each method under examination, in seconds. The first column of the table refers to the method name, while the next two columns present the time requirements of each technique to perform a feature drift assessment, using the \textit{fingerprinting} and the \textit{links} datasets correspondingly. Generally, the distance-based and statistics-based methods that exist in the literature are very fast, since they complete their assessments within $0.02$ to $0.4$ seconds. On the other hand, ALERT has larger execution times, ranging from $17.2$ - $3.5$ seconds, depending on the use case. This is expected, since ALERT partially trains an \gls{mlp} model, which severely affects its execution time.  Similarly, ML-based methods are slower compared to the others, since during the inference process they utilize a fully trained ML model. We should note that the execution times depicted in Table \ref{tab:exec-time} correspond to the inference process of the ML-based methods, and they do not account for the model training time requirements. We conclude that, in terms of absolute numbers, the execution time requirements of ALERT are affordable for real-world applications, even when large data volumes are involved.

Further,  gradual drifts usually occur in longer periods of time, from several hours, as noted by~\cite{costa2025analysis}, or several months, as seen in our \textit{fingerprinting} dataset. Considering this, our method's higher execution time can still be regarded as relatively fast, especially in the context of adapting to such evolving conditions.
\begin{figure}[t]
    \centering
    \includegraphics[width=0.35\textwidth]{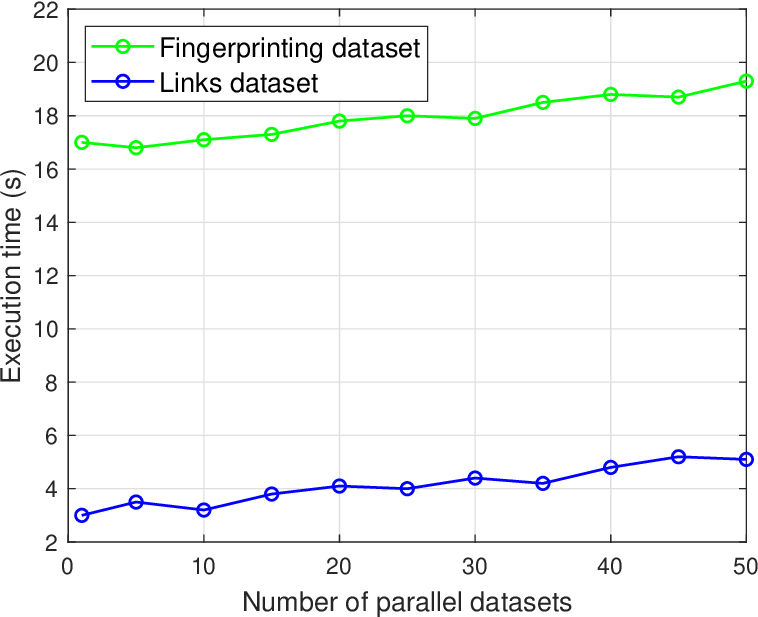}
    \caption {The execution time requirements of the ALERT methods when conducting several drift assessment tasks in parallel.}
    \label{exec_scale}
\end{figure}
ALERT's capacity to scale is illustrated in Figure \ref{exec_scale}, where we evaluate its performance when executing multiple drift assessments in parallel, simulating large-scale monitoring of diverse data sources. Each dataset contains 16,290 samples and is processed on a single machine with an NVIDIA GeForce GTX 1660 Ti (6 GB GPU) and an Intel i7-10750H (2.6 GHz CPU). The x-axis shows the number of datasets assessed in parallel, while the y-axis indicates the average execution time (in seconds) per assessment. The results show that ALERT scales efficiently, as the difference between 1 and 50 datasets is $\approx$2 seconds of execution overhead.

\begin{table*}[!htbp]
\caption{The details, advantages and disadvantages of the drift detectors under examination. }
\label{tab:drift_dif}
\resizebox{.99\textwidth}{!}{
  \begin{tabular}{c|cccc}
      & Statistical-based methods & Distance-based methods & ML-based methods & ALERT \\

\hline
Method type  & Unsupervised & Supervised & Unsupervised & Supervised\\

Parameters & p-value threshold & Probability bins, cut-off thresholds  & ML hyperparameters & {\gls{mlp}} hyperparameters, $T_s$    \\

Drift assessment  & Over data & Over data  & Over features & Over features  \\

Low parameter sensitivity   & \textcolor{red}{\XSolidBrush} & \textcolor{red}{\XSolidBrush} & \textcolor{blue}{\CheckmarkBold} & \textcolor{blue}{\CheckmarkBold}\\

Fast training  & \textcolor{blue}{\CheckmarkBold} & \textcolor{blue}{\CheckmarkBold} & \textcolor{red}{\XSolidBrush} & \textcolor{blue}{\CheckmarkBold}\\

Fast inference  & \textcolor{blue}{\CheckmarkBold} & \textcolor{blue}{\CheckmarkBold} & \textcolor{red}{\XSolidBrush} & \textcolor{red}{\XSolidBrush}\\

Resistant to noise  & \textcolor{red}{\XSolidBrush} & \textcolor{red}{\XSolidBrush} & \textcolor{blue}{\CheckmarkBold} & \textcolor{blue}{\CheckmarkBold}\\

Do not require regular re-training  & \textcolor{red}{\XSolidBrush} & \textcolor{red}{\XSolidBrush} & \textcolor{red}{\XSolidBrush} & \textcolor{blue}{\CheckmarkBold}\\

High accuracy with small datasets  & \textcolor{red}{\XSolidBrush} & \textcolor{red}{\XSolidBrush} & \textcolor{red}{\XSolidBrush}& \textcolor{blue}{\CheckmarkBold} \\

 \end{tabular}
}
\end{table*}

\subsection{Qualitative evaluation of drift detectors}

In this section, we discuss the drift detectors which are evaluated throughout this paper. We focus on breaking down their main advantages and disadvantages, and we present their input parameters. Our goal is to analyze the requirements of each technique and to assess their functionality in different scenarios. Under this premise, we also perform an analysis on how ALERT departs from the standard drift detection paradigm and what innovations it brings to the current discourse. Table \ref{tab:drift_dif} depicts several characteristics of the method groups under examination i.e., statistical-based, distance-based and ML-based.

Both statistical-based and distance-based methods conduct their drift assessments over the input data, are fast to train and support lightweight inference. On the other hand, they require frequent re-training in order to maintain high accuracy, they are not resistant to noise and they underperform when the training dataset is small. Further, they are very sensitive to input parameters which include the p-value threshold (for statistical techniques) and the number of probability bins, as well as the cut-off threshold, over which a drift is detected (for distance-based methods). A key difference between statistical-based and distance-based detectors is that the former are mostly unsupervised methods, while the latter usually fall into the category of supervised methods.

The ML-based techniques, including ALERT, utilise features (instead of raw data) to assess drifts, they have high resistance to noise and they are not very sensitive to their input parameters. On the flip side, their inference process is slower compared to statistical and distance based methods. The ML-based methods that exist in the literature require a significant amount of training time (ranging from minutes to hours), they can work with both labeled and unlabeled data, they need regular re-training to maintain high quality models, and their performance drops when the training dataset size is small. On the contrary, ALERT is a supervised method that does not require frequent re-training operations and works well with both large and smaller datasets.

\section{Conclusion}
\label{sec:conclusion}
In this paper we have introduced ALERT—a novel feature drift detection method that comprises  representation learning, statistical testing, and utility assessment components. We demonstrated ALERT's  superior performance on two real‑world wireless use cases, fingerprinting and link anomaly detection, where it outperformed ten established methods by ensuring that the AI models keep high F1‑scores, namely of 0.90 and 0.88 even in the presence of feature drift when it is suitably detected and re-training triggered.  Beyond raw detection accuracy, our work provides a comprehensive analysis workflow that not only pinpoints when and where feature drift occurs but also quantifies its impact on model performance and informs optimal retraining decisions. By rigorously benchmarking ALERT against standard and state-of-the-art approaches, we advance the state of the art in feature drift detection for wireless networks and offer practitioners a robust, end‑to‑end solution for maintaining reliable AI models in dynamic radio environments.

Although ALERT was evaluated on wireless network data, its architecture is domain-agnostic. Because it operates on learned feature representations rather than on specific radio-signal characteristics, the method can generalize to a wide range of time-series or tabular data domains. This includes applications such as financial transaction monitoring, healthcare sensor analysis, and industrial IoT systems, where data distributions evolve over time. Future work will explore adapting ALERT to these contexts, investigating how representation learning can capture domain-specific dynamics while maintaining robust drift detection across heterogeneous data sources. Additionally, a study of using such detection systems in large scale systems where feature dimensionality or incoming data volumes vary would also be relevant.

\section*{Acknowledgments}
This work was funded by the Slovenian Research and Innovation Agency (grant P2-0016) and by the European Union’s Horizon Europe Framework Program SNS-JU (grant agreement No. 101096456, NANCY). Instead of traditional spell-checking and language correction tools, some manually written paragraphs in the introduction and related work were refined using AI to improve flow and language quality.
\bibliographystyle{IEEEtran}
\bibliography{references.bib}

\section*{Biographies}
\vskip -2\baselineskip plus -1fil
\begin{IEEEbiography}
[{\includegraphics[width=1in,height=1.25in,clip,keepaspectratio]{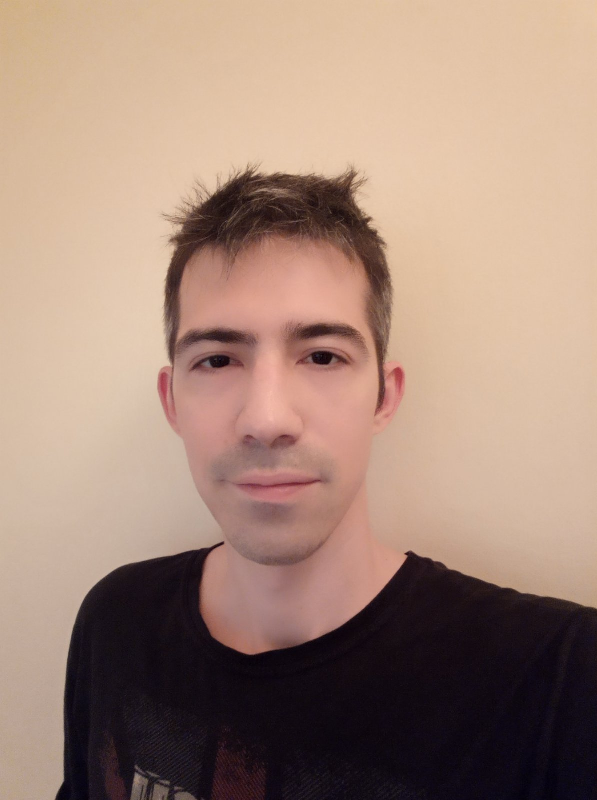}}]
{Athanasios Tziouvaras} received his B.Sc. and M.Sc. degrees in electrical engineering, and his Ph.D. in computer architecture and data-intensive applications from University of Thessaly in Greece. He joined Business and IoT integrated solutions Ltd. (BI2S) SME in 2021 and he is actively involved in research and innovation activities. His research interests include hardware acceleration, edge computing, machine learning, resource-aware computational methodologies and distributed computing. He has participated in more than 10 European and National research projects and has co-authored several publications in the domain of computer science.
\end{IEEEbiography}

\vskip -2\baselineskip plus -1fil
\begin{IEEEbiography}
[{\includegraphics[width=1in,height=1.25in,clip,keepaspectratio]{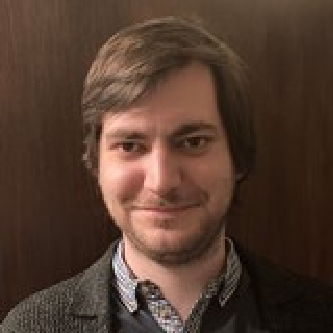}}]
{Bla\v{z} Bertalani\v{c}}
received his Ph.D. degree with highest distinction at the Faculty of Electrical engineering, University of Ljubljana. He is currently working as a researcher at Sensorlab, Jožef Stefan Institute. His main research interests are connected to advancement of machine learning and AI algorithms, especially in the context of time series analysis and smart infrastructures. Blaz is an IEEE member and with several leadership positions in the Slovenian chapter with over 15 IEEE publications.
\end{IEEEbiography}

\vskip -2\baselineskip plus -1fil
\begin{IEEEbiography}
[{\includegraphics[width=1in,height=1.25in,clip,keepaspectratio]{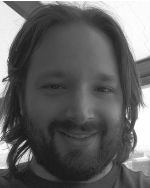}}]
{George Floros} is an Assistant Professor in the Department of Electronic \& Electrical Engineering at Trinity College Dublin. Previously, he was a Postdoctoral Researcher and Lecturer at the Department of Electrical and Computer Engineering at the University of Thessaly, Greece. He received his Diploma in Engineering, MSc, and PhD degrees from the same department in 2013, 2015, and 2019, respectively. His research interests primarily focus on the fields of electronic design automation (EDA), semiconductor device modeling, circuit simulation, model order reduction, thermal analysis of ICs, as well as VLSI design techniques and embedded systems. Throughout his academic career, he has authored over 10 journal articles, 30 conference papers, 3 posters, and abstracts.
\end{IEEEbiography}

\vskip -2\baselineskip plus -1fil
\begin{IEEEbiography}
[{\includegraphics[width=1in,height=1.25in,clip,keepaspectratio]{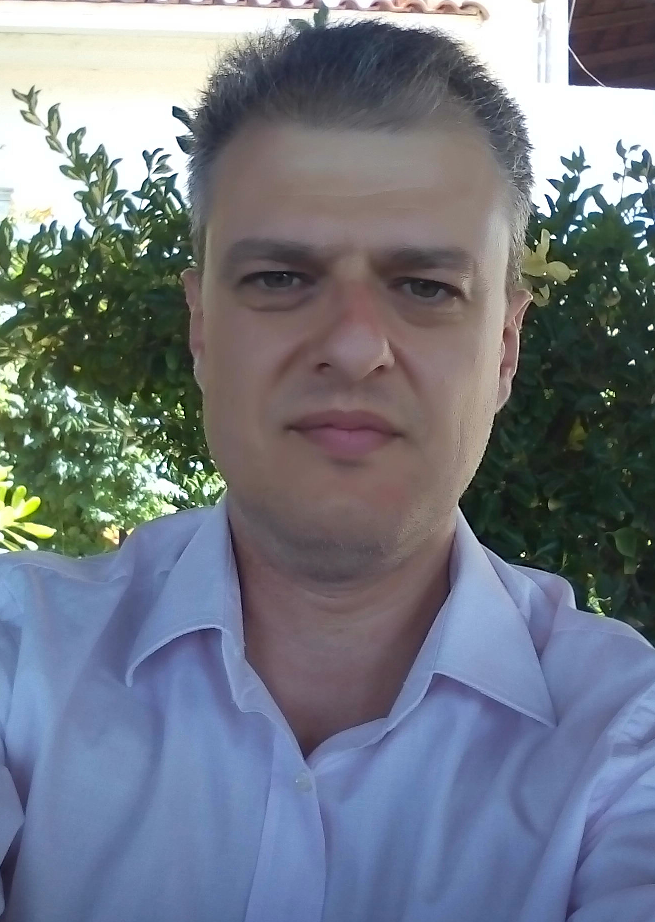}}]
{Kostas Kolomvatsos} received his B.Sc. in Informatics from the Department of Informatics at the Athens University of Economics and Business, his M.Sc. and his Ph.D. in Computer Science from the Department of Informatics and Telecommunications at the National and Kapodistrian University of Athens. Currently, he serves as an Assistant Professor in the Department of Informatics and Telecommunications, University of Thessaly. He was a Marie Skłodowska Curie Fellow (Individual Fellowship) at the School of Computing Science, University of Glasgow. His research interests are in the definition of Intelligent Systems adopting Machine Learning, Computational Intelligence and Soft Computing for Pervasive Computing, Distributed Systems, Internet of Things, Edge Computing and the management of Large-Scale Data. He is the author of over 130 publications in the aforementioned areas.
\end{IEEEbiography}

\vskip -2\baselineskip plus -1fil
\begin{IEEEbiography}
[{\includegraphics[width=1in,height=1.25in,clip,keepaspectratio]{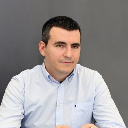}}]
{Panagiotis Sarigiannidis} received the B.Sc. and Ph.D. degrees in computer science from the Aristotle University of Thessaloniki, Greece, in 2001 and 2007, respectively. He is currently the Director of the ITHACA Laboratory, the Co-Founder of the 1st spin-off of the University of Western Macedonia, MetaMind Innovations, and a Full Professor with the Department of Electrical and Computer Engineering, University of Western Macedonia, in Greece. He has published more than 360 papers in international journals, conferences, and book chapters, including IEEE COMST, IEEE TRANSACTIONS ON COMMUNICATIONS, IEEE INTERNET OF THINGS, IEEE TRANSACTIONS ON BROADCASTING, IEEE SYSTEMS JOURNAL, IEEE ACCESS, and Computer Networks. He has been involved in several national, European, and international projects, including H2020 and Horizon Europe. His research interests include telecommunication networks, IoT, and network security. He received six best paper awards and the IEEE SMC TCHS Research and Innovation Award 2023. 
\end{IEEEbiography}

\vskip -2\baselineskip plus -1fil
\begin{IEEEbiography}
[{\includegraphics[width=1in,height=1.25in,clip,keepaspectratio]{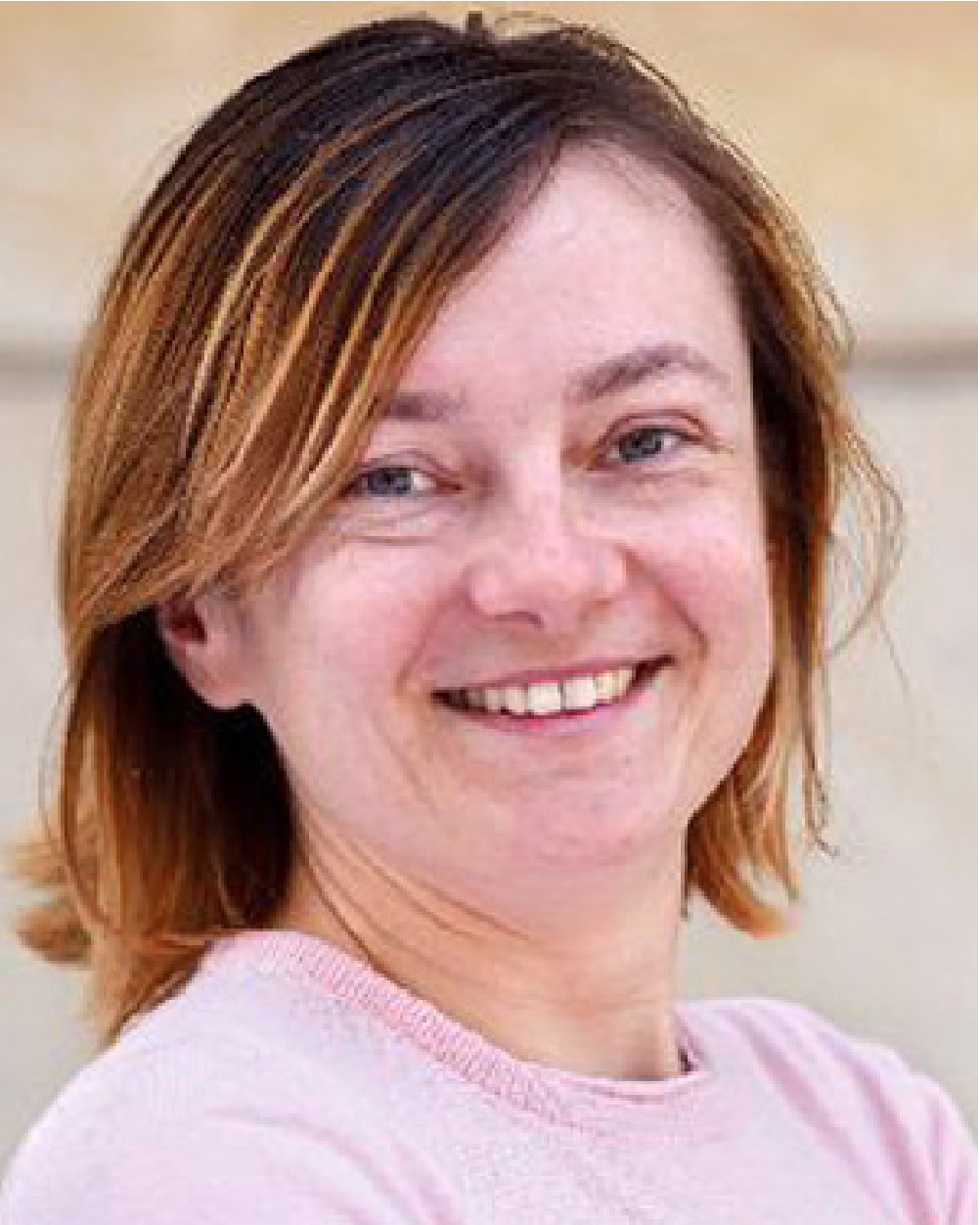}}]
{Carolina Fortuna} is a research associate professor  with the Jožef Stefan Institute where she leads Sensorlab. She was a Post-Doctoral Researcher with Ghent University, Ghent, Belgium. She was a Visiting researcher in Infolab with Stanford University, Stanford, CA, USA. She has led and contributed EU funded projects such as H2020 NRG5, eWINE, WISHFUL, FP7 CREW, Planetdata, ACTIVE, under various positions. She has advised/coadvised more than six M.Sc. and Ph.D. students. She has consulted public and private institutions. She has  coauthored over 100 papers including in IEEE COMST, IEEE WICOM Magazine, IEEE OJCOMS and Access. Her research interest includes developing the next generation smart infrastructures that surround us and improve the quality of our life. Dr. Fortuna contributed to community work as a TPC member, the track chair, and a TPC member at several IEEE conferences including Globecom and ICC.
\end{IEEEbiography}

\end{document}